\documentclass[conference]{IEEEtran}
\pdfoutput=1
\IEEEoverridecommandlockouts
\usepackage{cite}
\usepackage{amsmath,amssymb,amsfonts}
\usepackage{algorithmic}
\usepackage{graphicx}
\usepackage{textcomp}
\usepackage{xcolor}
\usepackage{tabularx}
\usepackage{booktabs}
\usepackage{url}
\usepackage{subcaption}
\usepackage{float}

\begin{document}

\pagestyle{plain}

\title{Explainability in AI-Based Applications – A Framework for Comparing Different Techniques\\
\thanks{* shared first authorship}
}

\makeatletter 
\newcommand{\linebreakand}{%
  \end{@IEEEauthorhalign}
  \hfill\mbox{}\par
  \mbox{}\hfill\begin{@IEEEauthorhalign}
}
\makeatother 

\author{
\IEEEauthorblockN{Arne Grobrügge*}
\IEEEauthorblockA{\textit{Karlsruhe Institute of Technology} \\
Karlsruhe, Germany \\
arne.grobruegge@outlook.de}
\and
\IEEEauthorblockN{Nidhi Mishra*}
\IEEEauthorblockA{\textit{Karlsruhe Institute of Technology} \\
Karlsruhe, Germany \\
nidhi.mishra@kit.edu}
\and
\IEEEauthorblockN{Johannes Jakubik}
\IEEEauthorblockA{\textit{Karlsruhe Institute of Technology} \\
Karlsruhe, Germany \\
johannes.jakubik@kit.edu}
\linebreakand

\IEEEauthorblockN{Gerhard Satzger}
\IEEEauthorblockA{\textit{Karlsruhe Institute of Technology} \\
Karlsruhe, Germany \\
gerhard.satzger@kit.edu}
}

\maketitle

\begin{abstract}

The integration of artificial intelligence (AI) into business processes has significantly enhanced decision-making capabilities across various industries such as finance, healthcare, and retail. However, explaining the decisions made by these AI systems poses a significant challenge due to the opaque nature of recent deep learning models, which typically function as black boxes. To address this opacity, a multitude of explainability techniques have emerged. However, in practical business applications, the challenge lies in selecting an appropriate explainability method that balances comprehensibility with accuracy. This paper addresses the practical need of understanding differences in the output of explainability techniques by proposing a novel method for the assessment of the agreement of different explainability techniques. Based on our proposed methods, we provide a comprehensive comparative analysis of six leading explainability techniques to help guiding the selection of such techniques in practice. Our proposed general-purpose method is evaluated on top of one of the most popular deep learning architectures, the Vision Transformer model, which is frequently employed in business applications. Notably, we propose a novel metric to measure the agreement of explainability techniques that can be interpreted visually. By providing a practical framework for understanding the agreement of diverse explainability techniques, our research aims to facilitate the broader integration of interpretable AI systems in business applications. 
\end{abstract}

\begin{IEEEkeywords}
Artificial Intelligence, Explainability, Agreement, Computer Vision Applications
\end{IEEEkeywords}

\section{Introduction}
In recent years, the rapid evolution of artificial intelligence (AI) has led to significant advancements in various sectors, including healthcare, finance, and autonomous systems. This development is primarily driven by the advent of deep learning technologies. 
These models are highly accurate in the processing of visual data and are increasingly leveraged across various application domains. 
%
Despite their powerful capabilities, most deep learning models, suffer from a critical drawback: their decision-making processes are largely opaque, making them difficult to interpret, especially in sectors where decisions have significant ethical or safety implications. This black box nature of AI systems raises substantial concerns in areas where accountability, transparency, and compliance with regulatory standards are essential \cite{castelvecchi2016can, adadi2018peeking}. For instance, in healthcare, accurate and interpretable diagnostics are crucial as they directly impact patient treatment plans and outcomes \cite{mondal2021xvitcos}. Similarly, in autonomous driving, understanding the basis for a vehicle’s navigational decisions is vital for safety and further development of autonomous technologies \cite{atakishiyev2021explainable}. 

The challenge of explainability in AI has led to the emergence of a specialized field known as Explainable Artificial Intelligence (XAI), which aims to make the outputs of AI systems more understandable to human users. XAI has become increasingly important as regulatory bodies, such as those in the European Union with the General Data Protection Regulation (GDPR), begin to require that automated decisions be explainable to the individuals they affect \cite{goodman2017european}. This regulatory landscape underscores a global shift towards greater transparency in AI, pushing the development of methods that can clarify the decision-making processes of complex deep learning models.
In response to these challenges, post-hoc explanation techniques \cite{slack2021reliable} have gained prominence within XAI. These methods seek to provide insights into the model's decisions after the model has been trained, without needing to modify the original model architecture. 
However, while numerous post-hoc explanation techniques exist, there is a scarcity of comprehensive research evaluating their effectiveness, particularly in the context of vision-based data.

This gap in research motivates our study, which aims to systematically evaluate and compare several leading post-hoc explanation techniques. We introduce a practical framework that involves generating and analyzing so-called attribution maps, which illustrate the impact of each pixel of the image on the model's prediction. By generating attribution maps across diverse explainability techniques, our proposed method allows for detailed qualitative and quantitative comparisons. With our analysis, we particular seek to address a fundamental question that remains largely unexplored in the current literature: How can we quantify the difference in the agreement of different post-hoc explanation techniques? 
We answer this question by presenting a detailed overview on the agreement of explanability techniques for one of the most popular deep learning architectures, the so-called Vision Transformer (ViT). Focusing on this architecture is particularly relevant as ViT allows the application of all available state-of-the-art post-hoc explanability techniques, and, therefore, supports the widest range of analyses. 
Overall, our objective is to offer understanding on the agreement of different explainability techniques to facilitate practical guidelines that can aid practitioners in selecting suitable explanation techniques for different applications. 


The paper is structured as follows: Section \ref{sec:Literature} reviews related work, establishing the significance and novelty of our approach. Section \ref{sec:Method} details our quantitative framework for evaluating explanation techniques, covering the generation, binarization, and effectiveness of attribution maps. Section \ref{sec:Setup} describes the experimental setup, including models, datasets, and explanation techniques. Section \ref{sec:Results} presents our findings on the performance of different explanation techniques across various metrics. We then discuss the implications, offering practical guidelines and future research directions. Finally, Section \ref{sec:Discussion} summarizes our contributions to explainable AI in real-world applications. 

\section{Literature}
\label{sec:Literature}
In this section, we review foundational deep learning concepts for computer vision to help familiarize less technical readers with key ideas relevant to our study. \cite{ali2023explainable} provide a thorough survey of Explainable AI (XAI), discussing the current landscape, challenges, and future directions for trustworthy AI systems. Their emphasis on the need for robust, transparent, and accountable AI models sets the context for our exploration of explainability methods in computer vision and Vision Transformers. We introduce key explainability techniques in these areas, followed by a discussion of various frameworks that aid decision-making in explainability and their applications in computer vision.


\subsection{Explainability in Computer Vision}

As the field of computer vision continues to advance, the demand for explainability in the underlying algorithms and models has become increasingly crucial.  Researchers have recognized the importance of not only achieving state-of-the-art performance but also providing users with a clear understanding of how these models arrive at their decisions \cite{escalante2018explainable}. This need for explainability extends beyond just the vision domain, as it is a fundamental requirement for building trust and ensuring the responsible deployment of these systems \cite{ras2022explainable}. Various methods have been developed to provide insights into the decision-making processes of these models. Among them, gradient-based, perturbation-based, and attention-based methods are widely used. While gradient-based and perturbation-based methods are general methods in computer vision, attention-based techniques are specific to the Transformer architecture and will be introduced in the subsequent subsection.

One key aspect of explainable computer vision is the ability to understand the internal representations learned by deep neural networks. Gradient-based models can provide insight into the importance of different features, highlighting the regions of the input that contribute the most to the model's predictions \cite{nielsen2022robust}. 
Similarly, perturbation-based methods, which involve systematically altering the input and observing the changes in the model's output, can help uncover the importance of different image regions \cite{ivanovs2021perturbation}. 

Despite these advancements, the field of explainable computer vision is still relatively young, and there are ongoing debates and challenges to be addressed. Some argue that fully explainable models may not be feasible for many complex tasks, as they often require a significant amount of domain knowledge and manual effort to design \cite{zablocki2022explainability}. This has lead to the proposal of diverse explanability techniques and we observe a void of meta methods that compare those different techniques. 

\subsection{Explainability in Vision Transformers}
The interpretability of deep learning models, particularly Vision Transformers (ViTs), is crucial for their adoption in sensitive and critical applications. Various methods have been proposed to enhance the explainability of ViTs, with attention-based methods being particularly significant due to their inherent use of attention mechanisms.

\cite{englebert2023explaining} introduced Transformer Input Sampling (TIS), a novel approach that leverages the flexibility of ViTs in processing variable numbers of input tokens to enhance model explainability without introducing misleading perturbations. Similarly, \cite{qiang2023interpretability} proposed an interpretability-aware training procedure that inherently enhances the interpretability of ViTs by prioritizing regions of interest, thereby improving model trustworthiness.

Attention-based methods exploit the attention weights within the transformer model to provide insights into the decision-making process. For instance, Attention Rollout aggregates attention weights across all layers of the transformer to visualize the cumulative attention distribution, helping to understand how information flows through the network and which regions are deemed important at different stages \cite{abnar2020quantifying, chefer2021transformer}. In addition to attention-based methods, gradient-based techniques like Grad-CAM have been adapted to visualize important regions in ViTs by using the gradients of the target concept to produce a localization map \cite{legrad2024}.

These advancements underscore the importance of developing models that not only perform well but also provide insights into their decision-making processes, thereby enhancing trust and reliability in high-stakes applications.

\subsection{Frameworks for Decision Support in Explainability Methods}
Frameworks that support decisions on explainability methods are essential in navigating the complex landscape of AI and machine learning, particularly in computer vision. These frameworks provide structured approaches for selecting suitable techniques to interpret and explain model decisions, considering factors like model type, application domain, and user requirements. IBM's AI Explainability 360 (AIX 360) offers a comprehensive library of algorithms and a taxonomy to help users understand and choose appropriate explainability techniques based on criteria such as transparency, interpretability, and fidelity \cite{arya2021ai}. Similarly, DARPA's XAI program focuses on creating explanations that are understandable to various user types, from AI developers to end-users \cite{gunning2019darpa}. User-centric frameworks like Interpretable Machine Learning (IML) emphasize tailoring explanations to the cognitive and contextual needs of different user groups \cite{doshi2017towards}. Additionally, application-specific frameworks like the Explainable AI Framework for Healthcare (XAI-H) ensure that explanations meet regulatory standards and support clinical decision-making processes \cite{holzinger2017we}. 

In their systematic review, \cite{vilone2020explainable} highlight the significant growth in XAI and cluster methods into four main categories, emphasizing the necessity for unified concepts and quantifiable metrics to tackle the opacity of complex models. Similarly, \cite{arrieta2020explainable} provide a comprehensive analysis of XAI literature, discussing the critical balance between model accuracy and interpretability and the importance of the human-in-the-loop. Both reviews underscore the ongoing challenges and opportunities in the field, aligning with our focus on integrating multiple explainability methods and developing standardized evaluation metrics to enhance the reliability and comprehensibility of AI systems in practical applications. Our paper specifically addresses the issue of integrating multiple explainability methods and developing standardized metrics by focusing on the agreement between different post-hoc explainability techniques. We have selected these challenges due to their practical relevance and the significant impact they can have on enhancing the reliability and comprehensibility of AI systems in real-world applications. By focusing on these challenges, we aim to contribute to the ongoing efforts to improve the trust, accountability, and usability of AI models across various domains.

\subsection{Applications of Computer Vision}
Computer vision, powered by advancements in deep learning and AI, has found extensive applications across a variety of fields, significantly enhancing the capabilities and efficiencies of numerous processes. These applications leverage the ability of computer vision models to analyze and interpret visual data, leading to innovations and improvements in sectors such as healthcare, autonomous driving, security, retail, and manufacturing. We shed light on particularly relevant applications, which are part of our experimental setup in the remainder of this work.

\subsubsection{Health}

In healthcare, computer vision is revolutionizing diagnostics and treatment planning. AI models can analyze medical images such as X-rays, MRIs, and CT scans to detect diseases at an early stage, often with greater accuracy than human practitioners. For example, Vision Transformers (ViTs) have been integrated into models for Alzheimer’s disease detection, providing interpretable features that improve diagnostic accuracy and aid in early intervention \cite{alzheimers_detection2024}. Similarly, explainable AI (XAI) techniques are being used in breast cancer detection, offering transparent outputs that help oncologists understand AI-driven insights and make better-informed decisions \cite{breast_cancer_detection2024}.

The technology also aids in automated health monitoring, where systems can analyze patient data to predict treatment outcomes and monitor disease progression. Applications include tracking vital signs and assessing surgical outcomes through real-time image analysis \cite{khanam2019remote}. Furthermore, computer vision systems are employed in medical training through simulation-based platforms, providing detailed feedback to trainees and enhancing their skills before actual surgeries.

\subsubsection{Manufacturing}
In manufacturing, computer vision is used for quality control and defect detection. AI models inspect products for defects, ensuring high quality and consistency on production lines. These systems can detect minute flaws that might be overlooked by human inspectors, thereby reducing waste and improving efficiency. Vision transformers have been applied to identify defects in products, such as wheels, with various explainability methods highlighting different defect aspects \cite{saberironaghi2023defect}.

\subsubsection{Environmental Monitoring}
Environmental monitoring benefits from computer vision through the analysis of satellite imagery to monitor deforestation, track wildlife, and assess environmental changes. These systems provide valuable insights into environmental health and changes over time, aiding in the development of effective conservation strategies \cite{bayoudh2022survey}.

\subsubsection{Autonomous Driving}
Autonomous vehicles rely heavily on computer vision to interpret data from cameras and sensors for navigation, obstacle detection, and real-time decision-making. These systems must be highly reliable and explainable to ensure safety and build public trust. By aligning machine perception with human visual attention, computer vision enhances the performance and safety of autonomous driving technologies \cite{atakishiyev2021explainable}.

The integration of computer vision in these diverse fields underscores its transformative potential. Explainable AI enhances the trustworthiness and reliability of these applications, ensuring that the insights provided by AI models are transparent and understandable. As technology evolves, computer vision applications will continue to expand, driving innovation and efficiency across various sectors.

To further illustrate the advancements in explainability methods for Vision Transformers, Table \ref{tab:xai_comparison} provides a comparative analysis of recent studies. This table highlights various frameworks, their objectives, methodologies, key findings, and unique aspects. By comparing these different frameworks, we can better understand the strengths and limitations of each approach, guiding future research and practical applications in the field.

\begin{table*}[ht]
\centering
\begin{tabular}{p{2cm}p{3cm}p{3cm}p{3cm}p{3cm}}
\toprule
\textbf{Reference} & \textbf{Objective} & \textbf{Methodology} & \textbf{Key Findings} & \textbf{Unique Aspects} \\ \midrule
AIX360 \cite{arya2021ai} & Comprehensive library of algorithms for explainability & Taxonomy for choosing techniques & Transparency, interpretability, fidelity & Broad algorithm library for general AI applications \\[0.3cm] \midrule
DARPA's XAI \cite{gunning2019darpa} & Understandable explanations for various user types & Creating user-understandable explanations & Importance of user types in understanding AI decisions & Focus on user-centric explanations \\[0.3cm] \midrule
IML \cite{doshi2017towards} & Tailored explanations to user needs & User-centric frameworks & Improved user trust and accountability & Emphasis on cognitive and contextual needs of users \\[0.3cm] \midrule
XAI-H \cite{holzinger2017we} & Explanations that meet healthcare regulatory standards & Application-specific frameworks tailored to healthcare & Enhanced clinical decision-making, regulatory compliance & Regulatory compliance in healthcare \\[0.3cm] \midrule
Faithfulness of ViT \cite{faithfulness_vit2024} & Assess faithfulness of post-hoc explanation techniques for ViTs & Evaluates various post-hoc explanation techniques & Strengths and limitations of XAI techniques & Focus on faithfulness of explanations \\[0.3cm] \midrule
XAI Personalization \cite{xai_personalization2024} & Improve recommendation systems using XAI and ViT & Combines XAI with clustering for recommendation systems & Enhances accuracy and personalization with explainable features & Content recommendations \\[0.3cm] \midrule
\textbf{This paper} & \textbf{Evaluate agreement among explainability techniques for ViTs} & \textbf{Quantitative framework for attribution maps} & \textbf{Significant disagreements among methods, use diverse methods} & \textbf{Focus on agreement of explainability techniques for ViTs, novel metric for measuring agreement} \\[0.3cm] 
\bottomrule
\end{tabular}
\vspace{5pt}
\caption{Comparative Analysis of Explainability Frameworks and Research}
\label{tab:xai_comparison}
\end{table*}

Overall, this section highlights the significant strides made in enhancing the explainability, efficiency, and domain-specific applicability of Vision Transformers. While substantial progress has been made, gaps still exist in understanding the full potential of ViTs, particularly in the context of enhancing the explainability of ViTs without compromising performance. Therefore, it is essential to holistically compare existing explainability techniques. Our work is mainly motivated by \cite{krishna2022disagreement}, who provide a quantitative framework that measures the disagreement between any two explanations. They coined the term "disagreement problem" to refer to this issue. Addressing the critique of attention-based explanation techniques by \cite{jain2019attention} and \cite{neely2021order}, we evaluated the paradigm of "agreement as evaluation," effectively comparing XAI methods using a Transformer architecture.

\section{Proposed Method}
\label{sec:Method}
In this paper, we propose a comprehensive framework for evaluating and comparing post-hoc explanation techniques applicable to Vision Transformer (ViT) models. 
Our approach systematically generates, binarizes, and evaluates attribution maps produced by various explanation techniques, facilitating a clearer understanding of the decision-making processes of ViT models.

\subsection{Generate Attribution Maps}
The first step involves creating what are known as attribution maps, which are essentially visual representations that highlight the specific areas within an image that influenced the model's prediction the most. These maps are crucial because they provide insights into what the model considers particularly important when making a decision. Depending on the explanation method used, these maps can be generated with respect to a particular class that the model is trying to predict. This means that the map highlights parts of the image that are most relevant to that class. Some explanation techniques produce maps that are class-specific, pinpointing exactly which features contribute to the prediction of a particular class, while others give a more general overview, without class specificity. The creation of these maps is a critical step in demystifying the model's predictions, making it possible to begin comparing how different explanation techniques perform. We particularly leverage two different types of attribution, represented by \textit{Integrated Gradients} and \textit{Perturbation-Based Attribution (LIME)}.

Integrated Gradients (IG) is a powerful technique for attributing the prediction of a neural network to its inputs. 
Given a model and an input, IG computes the gradients of the model output with respect to the input. The attribution is then obtained by integrating these gradients along the path from a baseline (typically a black image or the mean image) to the actual input.
In contrast, LIME offers an approach to generate explanations by locally approximating the model around the prediction. For a given prediction, LIME perturbs the input image and fits a simpler, interpretable model (e.g., a linear model) to the outputs obtained from these perturbations. 

\subsection{Binarization of Attribution Maps}
Given that the attribution maps contain detailed and often dense information, the next step simplifies these maps into binary versions. This process involves converting the detailed, colored, and gradient-filled maps into simple black and white images, where the white parts indicate regions of interest (positive contributions) and the black parts represent the background or non-contributory regions. This conversion is crucial for two reasons. Firstly, it allows for a more straightforward comparison between the maps generated by different explanation techniques. Secondly, it strips down the complexity of the maps, making them more accessible and easier to interpret for non-experts. This step involves setting all negative attribution values to zero, which helps in focusing solely on the regions positively contributing to the model’s decision. The process includes averaging, normalization, and dynamic thresholding techniques to ensure that the binary maps are consistently prepared across different explanation techniques.


\begin{figure}[h!]
    \includegraphics[width=0.48\textwidth]{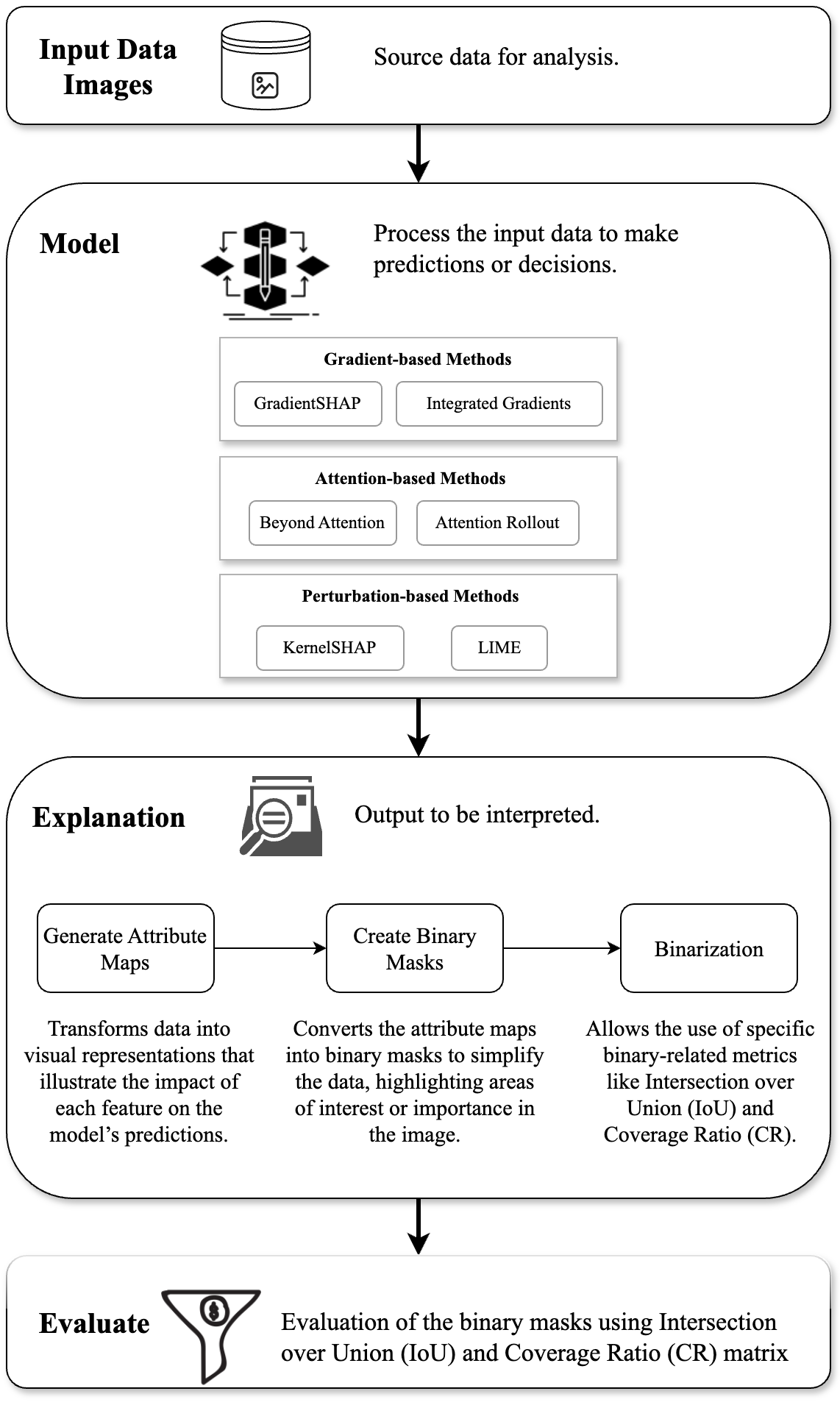}
    \caption{Decision Support Framework for the Evaluation of Different explainability techniques in Computer Vision Applications.}
    \label{fig:1}
\end{figure}

\subsection{Comparative Metrics}
The methodology incorporates quantitative metrics to evaluate and compare the binary masks derived from different explanation techniques, focusing on their ability to highlight the regions of an image that most significantly contribute to a model's prediction. Two primary metrics are employed for this purpose: Intersection over Union (IoU) and Coverage Ratio (CR). These metrics facilitate a systematic comparison by quantifying the degree of overlap and coverage between the binary masks generated by various explanation techniques.

\subsubsection*{Intersection over Union (IoU)}

The Intersection over Union (IoU) metric, also known as the Jaccard Index, is a common evaluation metric in the field of computer vision, particularly in tasks involving segmentation and object detection. It measures the overlap between two binary masks by calculating the ratio of their intersection area to their union area. Mathematically, the IoU between two binary masks, $mask_1$ and $mask_2$, is defined as:

\[
\text{IoU}(mask_1, mask_2) = \frac{|mask_1 \cap mask_2|}{|mask_1 \cup mask_2|}
\]

Where $|\cdot|$ denotes the area (i.e., the number of pixels) of the region, $\cap$ represents the intersection of the two masks, and $\cup$ denotes their union. The IoU score ranges from 0 to 1, where 0 indicates no overlap and 1 signifies perfect overlap between the two masks. This metric provides a straightforward and intuitive measure of similarity between the areas highlighted by different explanation techniques as being important for the model's decision.

\subsubsection*{Coverage Ratio (CR)}

While IoU offers valuable insights into the overlap between binary masks, it may not fully capture the nuances of coverage, especially when the masks vary significantly in size. To address this, the Coverage Ratio (CR) metric is introduced. CR quantifies the extent to which one mask covers another, providing a measure of coverage that is particularly informative in scenarios where one mask is much larger or more dispersed than the other. The Coverage Ratio of $mask_1$ with respect to $mask_2$ is defined as:

\[
\text{CR}(mask_1, mask_2) = \frac{|mask_1 \cap mask_2|}{|mask_1|}
\]

This equation calculates the ratio of the intersection area of the two masks to the area of $mask_1$, effectively measuring how much of $mask_1$ is covered by $mask_2$. Unlike IoU, CR is asymmetric, meaning that $\text{CR}(mask_1, mask_2)$ is not necessarily equal to $\text{CR}(mask_2, mask_1)$. This asymmetry allows for a more detailed examination of the relationships between the binary masks, shedding light on whether one explanation method tends to highlight broader or more specific regions of importance compared to another.

Together, the IoU and CR metrics provide a comprehensive framework for evaluating and comparing the binary masks derived from various post-hoc explanation techniques. By quantifying both the overlap and coverage between masks, these metrics enable a nuanced understanding of the similarities and differences in how each method interprets the contributions of different image regions to a model's predictions.

\subsection{Post-Processing of Binary Masks}
The final step acknowledges that different explanation techniques have varying granularities in how they assign importance across an image. For instance, gradient-based methods typically assign importance at the pixel level, creating highly detailed maps. In contrast, perturbation- and attention-based methods may operate on larger segments or patches of the image, leading to less granular maps. To ensure a fair and meaningful comparison, the method includes a post-processing step where binary masks generated by gradient-based methods undergo morphological operations to enhance their comparability with those from other types of methods. This includes techniques like morphological closing to unify scattered regions of importance, thereby aligning the level of detail across maps from different explanation techniques.

Through these steps, the proposed method offers a structured approach to comparing and evaluating different post-hoc explanation techniques for Vision Transformers. By generating, simplifying, and quantitatively comparing attribution maps, it provides insights into which methods offer more understandable and reliable explanations, guiding practitioners in selecting the most suitable tools for interpreting AI decisions in their specific contexts.

\section{Experimental Setup}
\label{sec:Setup}
This section outlines the experimental setup employed to evaluate the efficacy of various post-hoc explanation techniques applied to Vision Transformer (ViT) models in the context of image recognition. The setup is meticulously designed to ensure reproducibility of our experiments and to establish a clear framework for comparative analysis.

\subsection{Model Selection}
Our experiments utilize the Vision Transformer (ViT) model, specifically chosen due to its advanced performance in computer vision tasks. Originally proposed by Dosovitskiy et al. in 2020, the ViT model adapts transformer mechanisms, traditionally used in natural language processing, for image data processing. This model is particularly suitable for our study because it embodies the complexities typical of modern AI systems while providing a robust platform for testing explanation techniques. The model used in this study was pre-trained on the ImageNet-21k dataset and subsequently fine-tuned on ImageNet-1k, aligning with common practices in the field for achieving high accuracy in image classification tasks.

\subsection{Dataset Utilization}
For practical and computational efficiency, the experiments are conducted using the Imagenette dataset\footnote{\url{https://github.com/fastai/imagenette}}, a subset of ImageNet-1k. Imagenette includes ten easily classifiable classes from the broader ImageNet-1k dataset, which comprises a diverse range of images. This selection allows for focused analysis of the explanation techniques without the computational overhead of the full ImageNet dataset. The subset is chosen to test the robustness and effectiveness of the explanation techniques under controlled yet challenging conditions that mirror real-world scenarios.

\subsection{Explanation Method Configuration}
The experiments involve six state-of-the-art post-hoc explanation techniques: LIME, KernelSHAP, GradientSHAP, Integrated Gradients, Attention Rollout, and Beyond Attention. Each method is selected for its relevance and potential in providing insights into the decision-making processes of the ViT model:

\begin{itemize}
    \item \textit{LIME} and \textit{KernelSHAP} are perturbation-based methods that modify the input image in various ways to observe changes in output, thus identifying influential features.
    \item \textit{GradientSHAP} and \textit{Integrated Gradients} utilize gradients to determine the importance of each input feature, offering a direct measure of input-prediction relationships.
    \item \textit{Attention Rollout} and \textit{Beyond Attention} focus on the model's attention mechanisms, providing a unique perspective by highlighting how different areas of the input are weighted during the decision process.
\end{itemize}

\subsection{Computational Tools and Resources}
All explanation techniques are implemented using Python with the support of widely-used libraries such as PyTorch for model operations and Captum for gradient-based methods, ensuring high standards of reproducibility and accessibility. The choice of tools is intended to provide a balance between computational efficiency and the ability to conduct in-depth analyses. The experiments have been conducted on an NVIDIA A100 GPU.

\subsection{Open-Source Availability}
To contribute to the transparency and reproducibility of our research, we have made the source code available on GitHub\footnote{\url{https://github.com/grobruegge/ViTExplComp}}. This includes scripts for model training, explanation method implementation, and evaluation metrics calculation. By providing open access to our code, we encourage other researchers to replicate our experiments, compare results, and extend our work to other datasets or models.


\section{Results}
\label{sec:Results}

This section presents a detailed comparison of six post-hoc explanation techniques evaluated using the Vision Transformer model on the Imagenette dataset. We quantitatively and visually assess each method's efficacy in identifying influential regions of images.

\subsection{Quantitative Analysis}

We initiated our analysis by calculating the Intersection over Union (IoU) and Coverage Ratio (CR) for each pair of explanation techniques across various classes in the Imagenette dataset. The IoU metric gauges the overlap between binary masks generated by different methods, indicating methodological agreement in identifying key features. Conversely, the CR metric reveals the extent to which one method's identified regions are encompassed by another's, providing insight into the relative comprehensiveness or specificity of each method. Figures \ref{fig:2} and \ref{fig:3} display these metrics as colored heatmaps, averaged across all images. 

Our analyses reveal that the output of explainability techniques from the same methodological category (such as LIME and KernelSHAP) are characterized by a significant overlap. 
The CR scores, for example, show that 61 percent of the regions identified by LIME are also highlighted by KernelSHAP, indicating a significant agreement between methods from the same methodological group. In general, however, we observe a surprisingly low level of agreement among the XAI methods, with notable exceptions for pairs such as LIME and KernelSHAP, as well as Integrated Gradients (IG) and GradientSHAP, which demonstrate higher IoU scores due to their algorithmic similarities. This finding aligns with the observations made by \cite{krishna2022disagreement}, who noted significant disagreements among different explanation techniques. Such discrepancies highlight the necessity for using multiple complementary approaches to achieve more reliable and consistent explanations \cite{lakkaraju2016interpretable, ribeiro2016should}.

\begin{figure}[h!]
\centering
\includegraphics[width=0.45\textwidth]{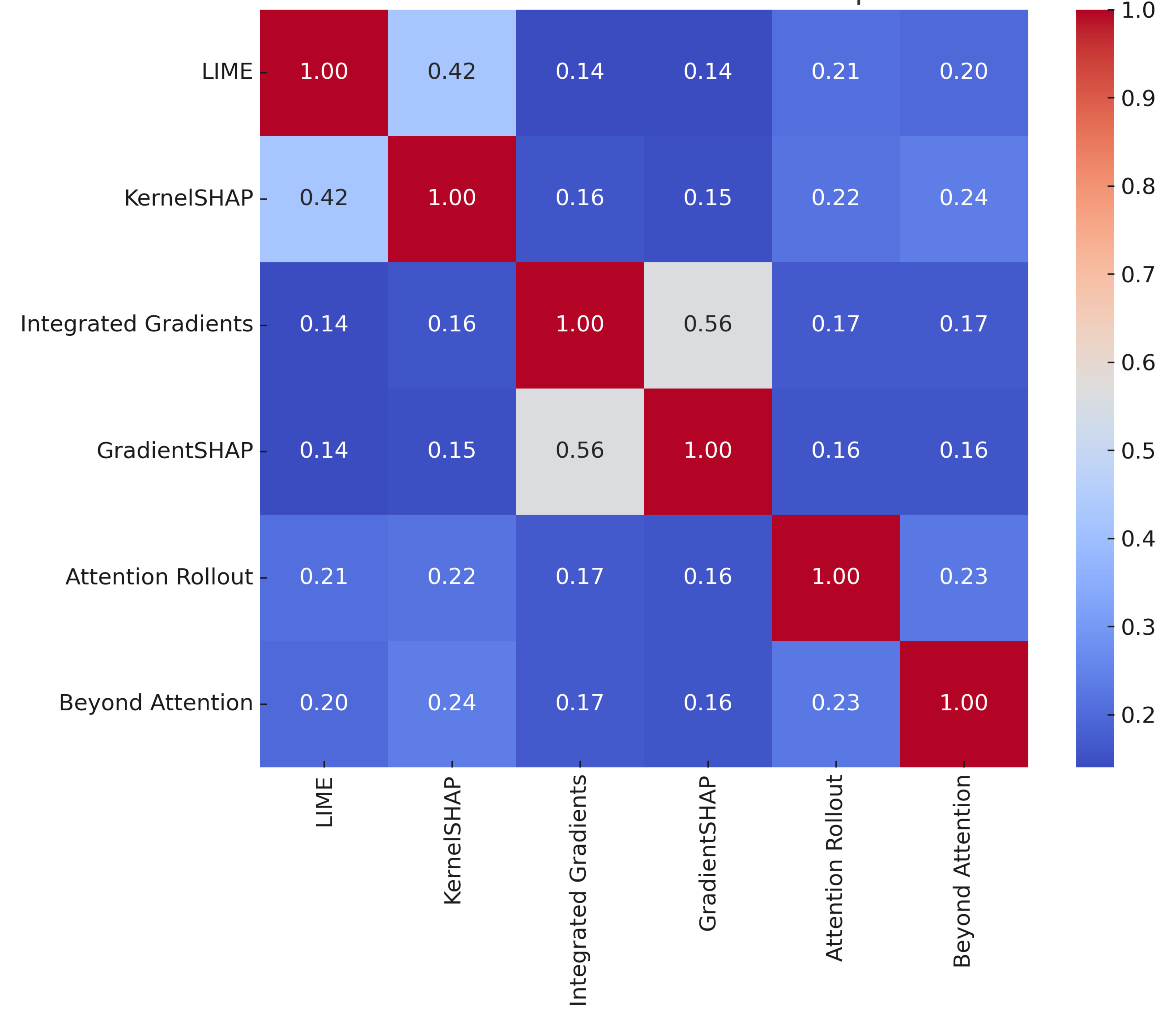}
\caption{Intersection over Union (IoU) scores showing agreement between different explainability techniques.}
\label{fig:2}
\end{figure}

\begin{figure}[h!]
\centering
\includegraphics[width=0.45\textwidth]{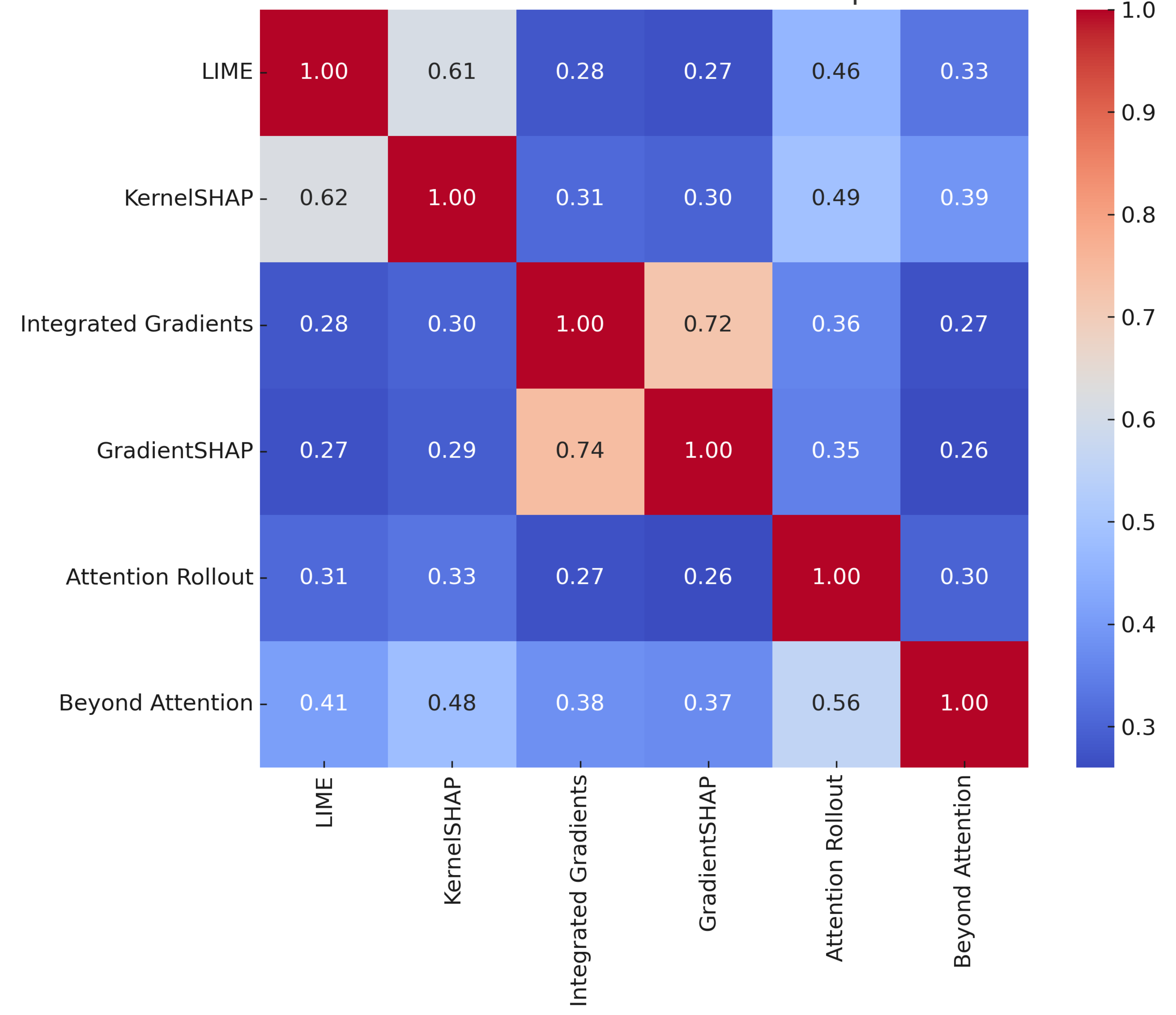}
\caption{Coverage Ratio (CR) scores reflecting the extent to which the attributions by one method are covered by those of another.}
\label{fig:3}
\end{figure}

Interestingly, while Attention Rollout tends to provide broader attribution maps, Beyond Attention consistently produces more focused, compact maps, as evidenced by its higher CR scores across all methods. This indicates that Beyond Attention's maps are often subsumed within those of other methods, particularly Attention Rollout. This is not only relevant as an isolated finding but demonstrates that our proposed evaluation metric CR allows in depth understanding of the differences between explainability methods that previously required manual visual inspection of the produced attribution maps. 

The IoU heatmap further reveals blocks of similar scores among methods with similar algorithms, highlighting the distinct clusters of explanation strategies: gradient-based, perturbation-based, and attention-based methods. This segmentation illustrates little correlation between gradient-based methods and other types, suggesting complementary strengths and weaknesses.

Overall, our findings suggest that employing a diverse array of explanation techniques can provide a more holistic understanding of model decisions, particularly when methods from different categories (gradient-based, perturbation-based, and attention-based) are used in tandem. This diversity allows for a richer interpretation of model behavior, accommodating both detailed and broad analysis needs depending on the specific requirements of the task at hand.

\subsection{Qualitative Analysis}

In addition to quantitative metrics that aggregate the explainability behavior of each method across the dataset, our proposed method allows for in-depth visual comparison of attribution maps. Figure~\ref{fig:attribution_maps} illustrates the attribution maps generated by several methods for three different use cases in the fields of sustainability, manufacturing, and healthcare. These visualizations help in understanding how each method emphasizes different aspects of the image when determining its class as well as the accuracy that can be expected by current explanation techniques.

\begin{figure*}[h]
    \centering
    \begin{subfigure}[b]{0.8\textwidth}
        \centering
        \includegraphics[width=\textwidth]{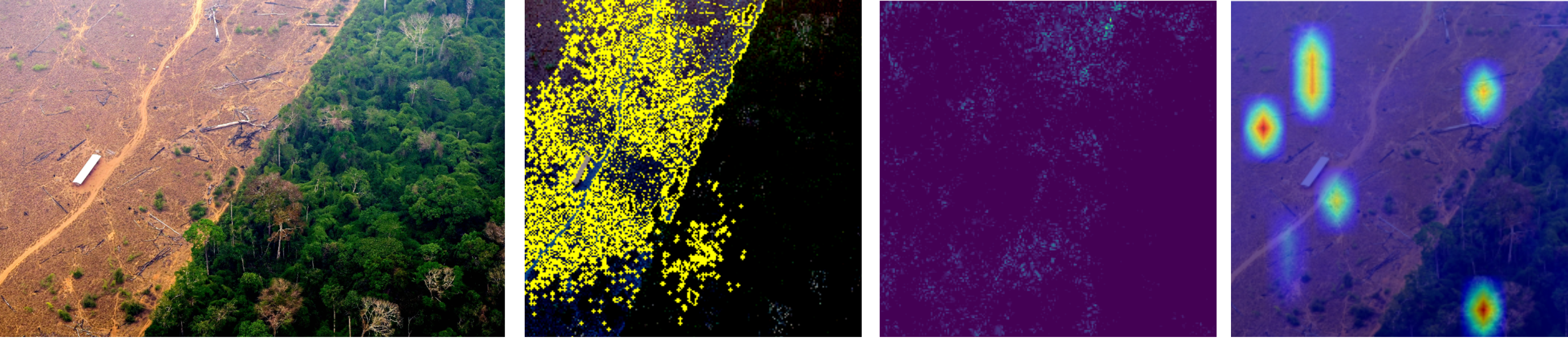}
        \caption{Environmental application: Measuring deforestation.}
        \label{fig:attribution_maps_a}
    \end{subfigure}\\\vspace{7pt}
    \begin{subfigure}[b]{0.8\textwidth}
        \centering
        \includegraphics[width=\textwidth]{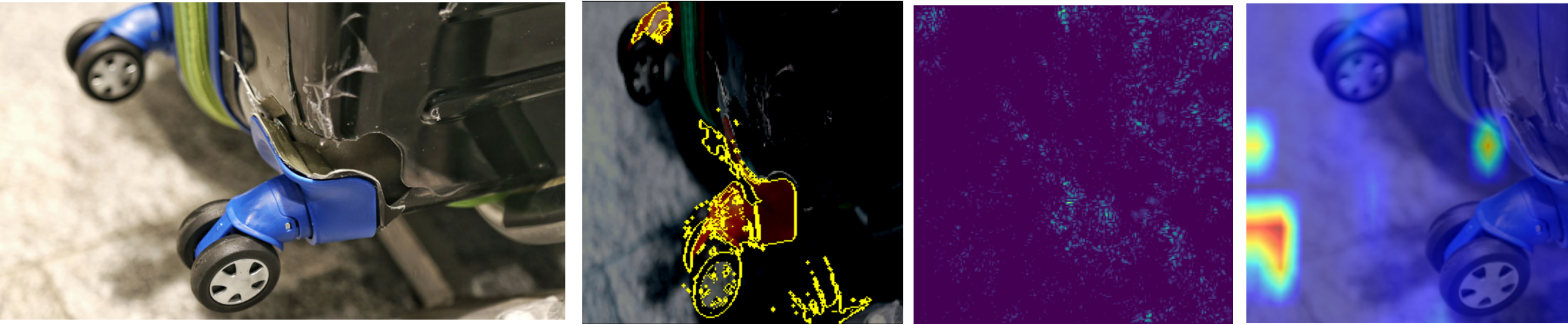}
        \caption{Manufacturing application: Defect identification.}
        \label{fig:attribution_maps_b}
    \end{subfigure}\\\vspace{7pt}
    \begin{subfigure}[b]{0.8\textwidth}
        \centering
        \includegraphics[width=\textwidth]{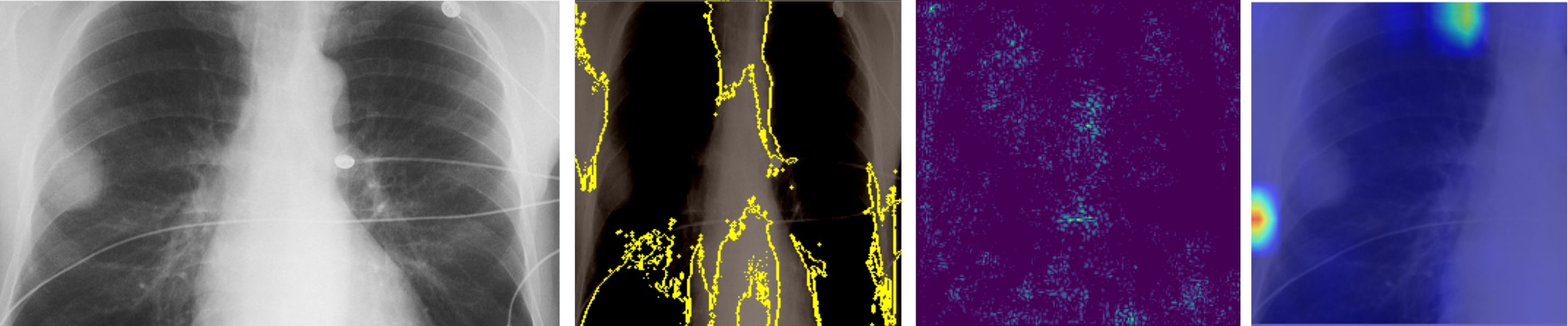}
        \caption{Healthcare application: Medical diagnoses.}
        \label{fig:attribution_maps_c}
    \end{subfigure}
    \caption{Differences across sample attribution maps generated by methodologically different explanation methods. From left to right: raw image, LIME-based explanation, integrated gradients-based explanation, and attention rollout-based explanation.}
    \label{fig:attribution_maps}
\end{figure*}

Figure~\ref{fig:attribution_maps} showcases the differences across sample attribution maps generated by various post-hoc explanation methods—LIME, Integrated Gradients, and Attention Rollout—applied to different applications. In the environmental application (a), the attribution maps highlight the areas most indicative of deforestation, with each method focusing on different features of the landscape. In the manufacturing application (b), the methods identify key regions indicating a defect in a product, such as the wheel, with varying levels of precision and area coverage. For the healthcare application (c), the attribution maps for medical diagnoses reveal the significant regions within a chest X-ray, with each method spotlighting different parts of the lungs and potential anomalies. These visual differences underline the importance of selecting appropriate explanation techniques for specific tasks to ensure accurate and reliable interpretations. For example, in the case of deforestation, the examination of whether the model's decision-making process was correct may be inconclusive for attention rollout, while it is more clear from the LIME-based attribution maps.


\section{Discussion}
\label{sec:Discussion}

This study proposes a practical framework to assess the agreement of different explainability methods, an area of research that is essential yet underexplored compared to applications in textual or tabular data. The research contributes to the literature by developing both a general-purpose quantitative framework and a corresponding metric to evaluate and compare various post-hoc explanation techniques tailored for Vision Transformers.
Based on our results, we derive the following practical implications which we summarize in Figure~\ref{fig:results}. 

\begin{figure}[h!]
\centering
\includegraphics[width=0.45\textwidth]{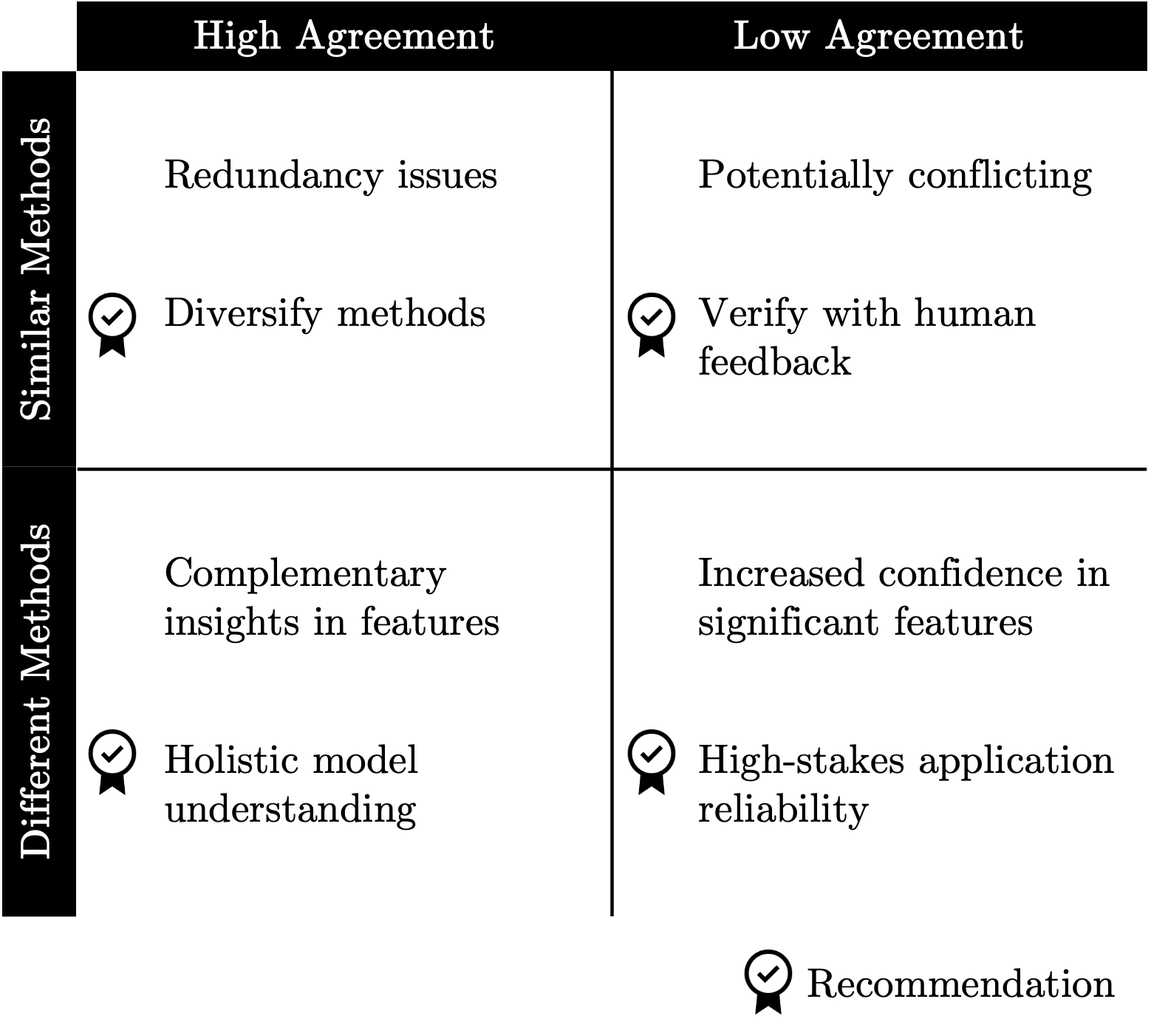}
\caption{Practical implications of agreement levels in explainability techniques for Vision Transformers, categorized by high/low agreement and similar/different methods.}
\label{fig:results}
\end{figure}

First, the high agreement of methodologically similar explainability techniques can lead to redundancy. For instance, LIME and KernelSHAP, as well as Integrated Gradients and GradientSHAP, show significant overlap in their explanations due to their similar underlying algorithms. Practitioners should be aware of this redundancy and may choose to avoid using multiple similar methods in favor of diversifying the techniques employed to save on computational resources and reduce unnecessary complexity.
Second, similar explainability techniques with low agreement are potentially conflicting as, based on our experiments, we would expect a high agreement of similar techniques. When, in contrast, we observe a low agreement, human assessment of the actual explanation outputs is necessary to understand whether one or multiple methods fail on specific data instances. 
Third, when different explainability techniques show a high agreement, they likely highlight the \textit{same} aspects of the \textit{same} underlying concept. Techniques that methodologically diverge significantly, such as gradient-based and perturbation-based techniques (see Figure~\ref{fig:attribution_maps}), can still provide provide similar outputs that, when combined, offer a more holistic understanding of the model’s behavior. By integrating these diverse explanations, practitioners can uncover particularly influential features that are highlighted by multiple explanation techniques and correspondingly identify potential model weaknesses.
Fourth, the low agreement of methodologically different approaches can be particularly relevant to explain the decision of a deep learning model. When different types of methods, such as gradient-based and attention-based techniques, diverge, they highlight \textit{different} aspects of the \textit{same} underlying concept. This can help to increase confidence in the identified features being truly significant. This cross-method agreement can be crucial in high-stakes applications, where validating the reliability of explanations is paramount for trust and accountability. 

However, it is important to address the limitations and criticisms of some popular post-hoc explanation methods such as \cite{slack2020fooling} have shown that these methods can be vulnerable to adversarial attacks, which can significantly undermine their reliability. Such vulnerabilities mean that these methods can be manipulated to produce misleading explanations, raising concerns about their robustness in critical applications. Furthermore, \cite{mittelstadt2019explaining} critique the overreliance on post-hoc explanations, noting that humans may misinterpret these explanations or place undue trust in them. This highlights the necessity for caution and critical evaluation when using these methods, and underscores the importance of using multiple explanation techniques to cross-verify results and mitigate potential biases.

Overall, utilizing several post-hoc explanation techniques can aid the trustworthiness of AI models by identifying underlying similarities and discrepancies. This approach mitigates the inherent black-box nature of AI systems \cite{samek2021explaining, dovsilovic2018explainable, vilone2020explainable, dosovitskiy2020image}. By comparing different methods, practitioners can discern consistent patterns, leading to a more reliable understanding of the model's decision-making process. This multi-method approach not only provides a safety net by cross-verifying results but also helps in uncovering any potential biases or systematic errors present in individual methods.



Our results underscore the importance of methodological diversity in explainability. The low overall agreement among methods suggests that no single method is universally superior. However, pairs of methods with algorithmic similarities, such as LIME \& KernelSHAP and Integrated Gradients \& GradientSHAP, demonstrate higher agreement. Attention-based methods like Beyond Attention and Attention Rollout show unique strengths, with Beyond Attention providing more compact maps and Attention Rollout offering more comprehensive coverage. To achieve a balanced and thorough interpretation of AI models, practitioners should consider using at least one method from each methodological category (i.e., gradient-based, perturbation-based, attention-based). Based on the visual comparison of the attribution maps, our results show that gradient-based methods, such as Integrated Gradients and GradientSHAP, offer detailed, pixel-level insights into how each part of the input contributes to the model's output. Perturbation-based techniques, like LIME and KernelSHAP, are intuitive and model-agnostic \cite{ribeiro2016model}, providing straightforward explanations by perturbing input data and observing changes in predictions. Attention-based methods, such as Attention Rollout and Beyond Attention, focus on the internal attention mechanisms of transformer models, explaining which parts of the input the model focuses on most.

\section{Conclusion}

Measuring the agreement of explanation techniques provides insights provide practical guidelines for selecting suitable post-hoc explanation techniques, thereby enhancing the transparency, trustworthiness, and overall acceptance of AI systems in critical applications. We are looking forward to see how researchers and practioners make use of our method in future research to expand our analyses to other deep learning architectures and datasets, further refining the guidelines for method selection and application-specific requirements.

Future work should address the challenge of evaluating which explainability or interpretability method is best suited for a given context, considering the model, dataset, task, and user background knowledge. As our study shows, different methods can highlight various aspects of the model’s behavior, and the choice of method may significantly impact the interpretability and trustworthiness of the AI system. \cite{miller2023explainable} emphasizes that explainability is a highly contextual task, and developing hypothesis-driven decision support using evaluative AI could enhance the effectiveness of explainability techniques across diverse applications. This approach would involve tailoring the explanation methods to specific use cases, ensuring that the chosen techniques provide the most relevant and actionable insights for the intended users (Miller, 2023).

We look forward to seeing how researchers and practitioners make use of our method in future research to expand our analyses to other deep learning architectures and datasets. This will further refine the guidelines for method selection and application-specific requirements, ultimately contributing to the development of more transparent and trustworthy AI systems.






\bibliographystyle{unsrt}
\bibliography{bib}

\end{document}